\documentclass[journal]{IEEEtran}
\usepackage[numbers,sort&compress]{natbib}
\usepackage{graphicx}
\usepackage{amsmath}
\usepackage{amssymb}
\usepackage{multirow}
\usepackage{mathrsfs}
\usepackage{amssymb}
\usepackage{pifont}
\newcommand{\cmark}{\ding{51}}%
\newcommand{\xmark}{\ding{55}}%
\usepackage{color}
\definecolor{dgreen}{rgb}{0.187,0.580,0.255}
\definecolor{purple}{RGB}{81,35,115}

\usepackage[pagebackref=false,breaklinks=true,letterpaper=true,colorlinks,citecolor=blue,linkcolor=blue,bookmarks=false]{hyperref}

\begin{document}
%
\title{SINet: A Scale-insensitive Convolutional \\ Neural Network for Fast Vehicle Detection}
%
%
%

\author{Xiaowei Hu, ~\IEEEmembership{Student Member,~IEEE,}
        Xuemiao Xu, ~\IEEEmembership{Member,~IEEE,}
        
        Yongjie Xiao, 
        Hao Chen, ~\IEEEmembership{Student Member,~IEEE,}
        Shengfeng He, ~\IEEEmembership{Member,~IEEE,}
        
        Jing Qin, ~\IEEEmembership{Member,~IEEE,}
        and Pheng-Ann Heng,~\IEEEmembership{Senior Member,~IEEE}

\thanks{Manuscript received July 14, 2017; revised January 1, 2018 and April 2, 2018; accepted May 9, 2018.}
\thanks{Corresponding author: Xuemiao Xu (xuemx@scut.edu.cn).}
\thanks{X. Hu started this work when he was an undergraduate student at the School of Computer Science and Engineering, South China University of Technology, and he finished it when he has become a Ph.D. student at the Department of Computer Science and Engineering, The Chinese University of Hong Kong.}
\thanks{X. Xu, Y. Xiao and S. He are with the School of Computer Science and Engineering, South China University of Technology.}
\thanks{H. Chen is with the Department of Computer Science and Engineering, The Chinese University of Hong Kong.}
\thanks{J. Qin is with Centre for Smart Health, School of Nursing, The Hong Kong Polytechnic University.}
\thanks{P.-A. Heng is with the Department of Computer Science and Engineering, The Chinese University of Hong Kong and Guangdong Provincial Key Laboratory of Computer Vision and Virtual Reality Technology, Shenzhen Institutes of Advanced Technology, Chinese Academy of Sciences, China.}

}

%
%


\markboth{IEEE TRANSACTIONS ON INTELLIGENT TRANSPORTATION SYSTEMS}
{Shell \MakeLowercase{\textit{et al.}}: Bare Demo of IEEEtran.cls for IEEE Journals}

%



\maketitle

\begin{abstract}
Vision-based vehicle detection approaches achieve incredible success in recent years with the development of deep convolutional neural network (CNN).
However, existing CNN-based algorithms suffer from the problem that the convolutional features are scale-sensitive in object detection task but it is common that traffic images and videos contain vehicles with a large variance of scales.
In this paper, we delve into the source of scale sensitivity, and reveal two key issues: 1) existing RoI pooling destroys the structure of small scale objects; 2) the large intra-class distance for a large variance of scales exceeds the representation capability of a single network.
Based on these findings, we present a scale-insensitive convolutional neural network (SINet) for fast detecting vehicles with a large variance of scales. First, we present a context-aware RoI pooling to maintain the contextual information and original structure of small scale objects. Second, we present a multi-branch decision network to minimize the intra-class distance of features.
These lightweight techniques bring zero extra time complexity but prominent detection accuracy improvement.
The proposed techniques can be equipped with any deep network architectures and keep them trained end-to-end.
Our SINet achieves state-of-the-art performance in terms of accuracy and speed (up to 37 FPS) on the KITTI benchmark and a new highway dataset, which contains a large variance of scales and extremely small objects.
\end{abstract}


%
\IEEEpeerreviewmaketitle

\section{Introduction}

\begin{figure} [t]
	\begin{center}

		\includegraphics[width=1.0\linewidth]{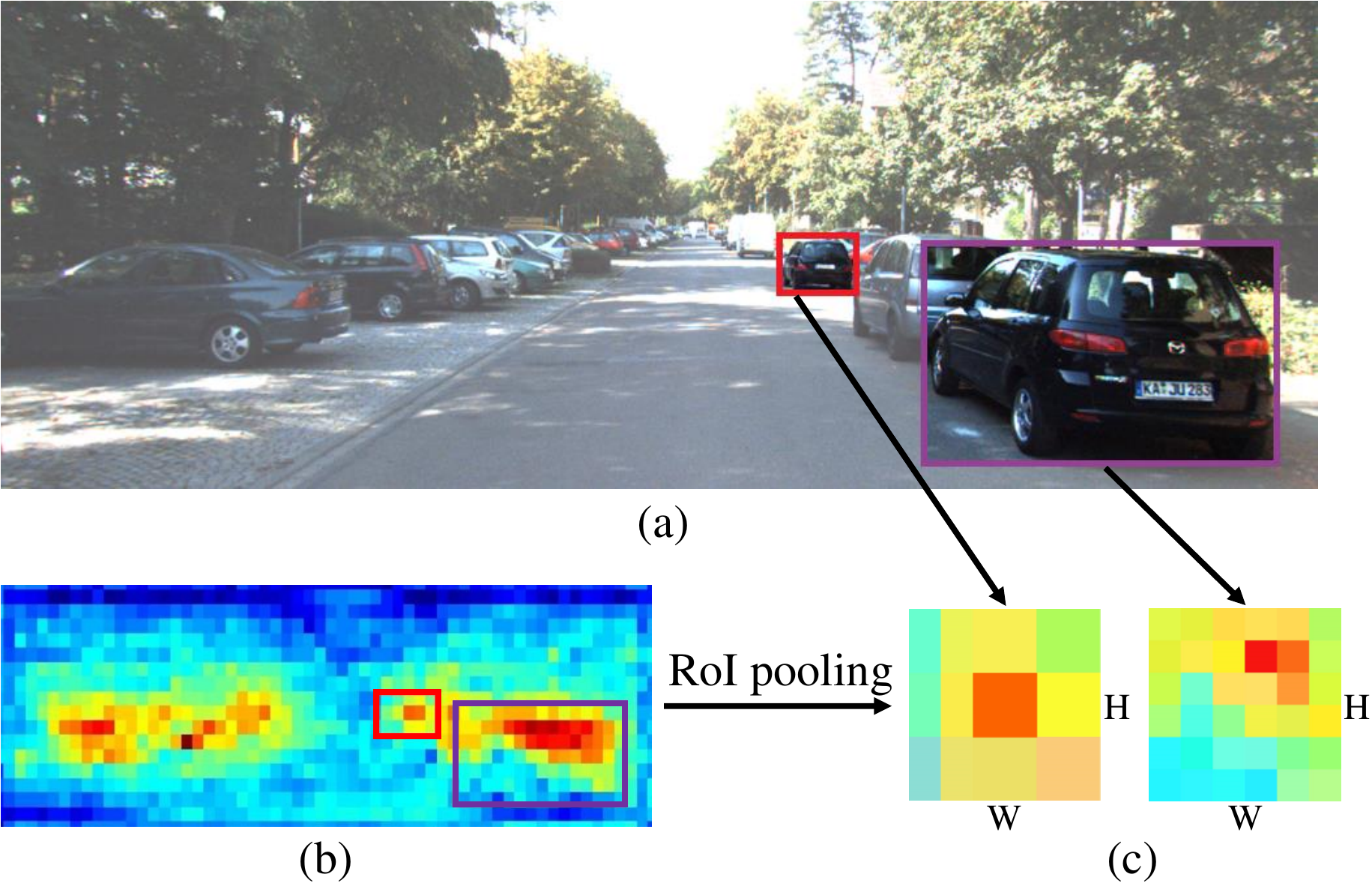}
	\end{center}
    \vspace{-3mm}
	\caption{The scale-sensitive problem. (a) An image includes both large and small vehicles. (b) The feature representations of small and large vehicles in deep layer are largely different. (c) Traditional RoI pooling introduces noise as it simply replicates the values on the feature map for the small vehicle.}
	\label{fig:motivation}
	\vspace{-5mm}
\end{figure}

\begin{figure*} [ht]
	\begin{center}
		\includegraphics[width=1.0\linewidth]{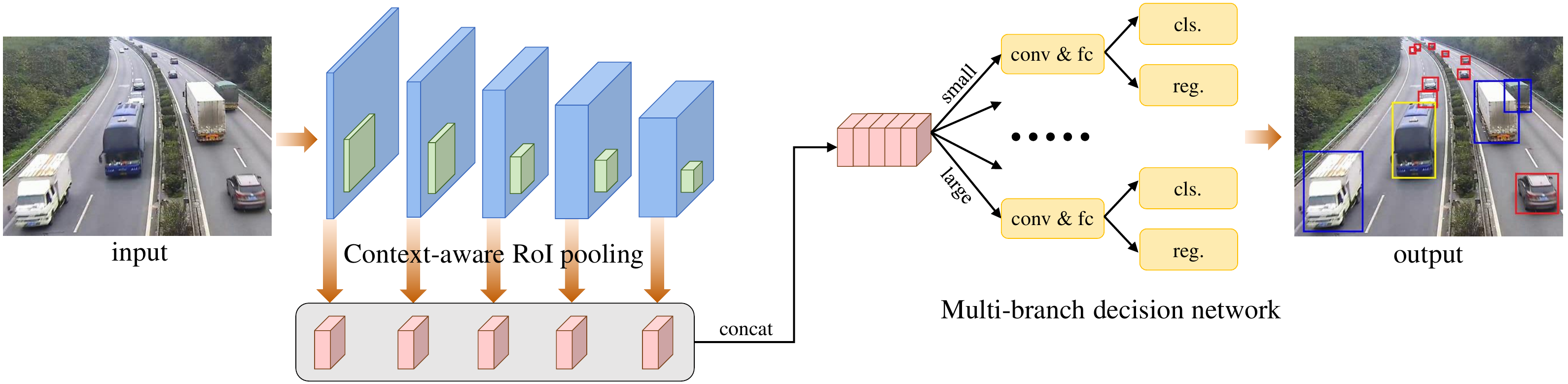}
	\end{center}
    \vspace{-3mm}
	\caption{
	The schematic illustration of the pipeline of the proposed SINet: (i) we extract feature maps with multiple scales over the CNN~\cite{lecun2015deep} from the input image and get the proposals based on the CNN features~\cite{cai2016unified}; (ii) each proposal on different layers is pooled into a fixed-size feature vector using the  context-aware RoI pooling, in which the small proposals are enlarged by the deconvolution with bilinear kernels to achieve better representation (see section~\ref{sec:CARoIpooling} for details); (iii) we concatenate the features of proposals at each layer and feed them to the multi-branch decision network; and (iv) lastly, we fuse the predicted bounding boxes from all branches to produce the final detection results (car in red, bus in yellow, van in blue). Best viewed in color.}

	\label{fig:arc}
	\vspace{-3mm}
\end{figure*}

\IEEEPARstart{A}{utomatic} vehicle detection from images or videos is an essential prerequisite for many intelligent transportation systems.
For example, vehicle detection from in-car videos (Fig.~\ref{fig:motivation}) is critical for the development of autonomous driving systems while vehicle detection from surveillance videos (Fig.~\ref{fig:arc}) is fundamental for the implementation of intelligent traffic management systems.
In this regard, over the past decade, a lot of effort has been dedicated to this field~\cite{li2014integrating,wu2016learning,stauffer1999adaptive,chen2012vehicle,premebida2006multi,martinez2008driving,cui2010vehicle,kyo1999robust,vargas2010enhanced,sun2004road,yuan2011learning,niknejad2012road,sun2006monocular,hsieh2014symmetrical,chang2010online,sivaraman2010general,yebes2014supervised,wang2017joint,yuan2017incremental,wang2018embedding}.
Some challenging benchmarks have also been proposed for evaluation and comparison of various detection algorithms~\cite{geiger2012we}.
On the other hand, in recent years, deep convolutional neural networks (CNNs) have achieved incredible success on vehicle detections as well as various other object detection tasks~\cite{bell2016inside,girshick2015fast,girshick2014rich,he2015spatial,he2016deep,ren2015faster,cai2016unified,xiang2017subcategory,yang2016exploit}.
However, when applying CNNs to vehicle detection, one of the main challenges is that traditional CNNs are sensitive to scales while it is quite common that in-car videos or transportation surveillance videos contain vehicles with a large variance of scales (see the vehicles in Fig.~\ref{fig:motivation} (a) and the input of Fig.~\ref{fig:arc}).
The underlying reason of this scale-sensitive problem is that it is challenging for a CNN to response to all scales with optimal confidences~\cite{sermanet2013overfeat}.

Existing CNN-based object detection algorithms attempt to make the network fit different scales by utilizing input images with multiple resolutions~\cite{girshick2015fast,girshick2014rich,he2016deep,sermanet2013overfeat,shrivastava2016training,xiang2017subcategory,xie2016aggregated} or fusing multi-scale feature maps of CNN~\cite{bell2016inside,cai2016unified,he2015spatial,kong2016hypernet,lin2016feature,liu2016ssd,shelhamer2017fully,shrivastava2016beyond,yang2016exploit,zagoruyko2016multipath}.
These methods, however, introduce expensive computational overhead and thus are still incapable of fast vehicle detection, which is essential for autonomous driving systems, real-time surveillance and prediction systems.

Instead of simply adding extra operations, we look into the detection network itself and scrutinize the underlying reasons of this scale-sensitive problem.
We observe two main barriers.
First, inadequate and/or imprecise features of small regions lead to the loss of detecting small objects (e.g., the red box in Fig.~\ref{fig:motivation} (b)).
In particular, the commonly used RoI pooling~\cite{girshick2015fast} distorts the original structure of small objects, as it simply replicates the feature values to fit the preset feature length (as shown in the left example of Fig.~\ref{fig:motivation} (c)). Second, the intra-class distance between different scales of vehicles is usually quite large.
As illustrated in Fig.~\ref{fig:motivation} (b), the red and purple boxes have different feature responds.
This makes it difficult for the network to represent objects with different sizes using the same set of weights.

To cope with the above problems, we present a scale-insensitive convolutional neural network, named SINet, to detect vehicles with a large variance of scales accurately and efficiently.
The network architecture is shown in Fig.~\ref{fig:arc}.
Object proposals are used on the feature map level to examine all the possible object regions, and the corresponding feature maps are fed to a decision network.
Two new methods are proposed to overcome above-mentioned barriers.
We first present a context-aware RoI pooling scheme to preserve the original structures of small scale objects.
This new pooling layer involves a deconvolution with bilinear kernels which can maintain the context information and hence help produce features that are faithful to the original structure.
These pooled features are then fed to a new, multi-branched decision network.
Each branch is designed to minimize the intra-class distance of features, and therefore the network can more effectively capture the discriminative features of objects with various scales than traditional networks.

The proposed network achieves state-of-the-art performance on both detection accuracy and speed on the KITTI benchmark~\cite{geiger2012we}.
This method also shows a promising performance on detecting vehicles with low resolution input images, and brings detecting vehicles in low-resolution video surveillance into practice.
Due to the lightweight architecture, real-time detection (up to $37$ FPS) can be achieved on a 256$\times$846 image.
In order to demonstrate the proposed method in more practical scenes, we construct a new highway dataset, which contains vehicles with a vast variance of scales.
To the best of our knowledge, it is the first dataset focuses on the highway scene.
It contains $14388$ well labelled images under different roads, time, weathers and traffic states.
This dataset, as well as the source code of the SINet, are publicly available at \url{https://xw-hu.github.io/}.
In summary, our contributions include:

\begin{itemize}
	\item[-]We present a context-aware RoI pooling layer, which can produce accurate feature maps for vehicles with small scales without extra space and time burdens. The proposed new pooling layer can be widely applied to existing architectures.
	\item[-]We present a multi-branch decision network for vehicle detection. It can accurately classify vehicles with a large variance of scales without introducing extra computational cost.
	\item[-]We construct the first large scale variance highway dataset, which provides a platform with practical scenes to evaluate the performance of various vehicle detection algorithms in handling target object with a large variance of scales.
	
\end{itemize}

\section{Related Works on Vehicle Detection}

In this section, we give a brief introduction of the monocular vision vehicle detection methods, as our approach also belongs to the monocular vision detection.
More comprehensive analysis of vehicle detection on monocular, stereo, and other vision-sensors can be found in~\cite{sivaraman2013looking}.

Early works use the relative motion cues between the objects and background to detect the vehicles. Adaptive background models such as Gaussian Mixture Model (GMM)~\cite{stauffer1999adaptive, chen2012vehicle, premebida2006multi}, Sigma-Delta Model~\cite{vargas2010enhanced} are widely used in vehicle detection by modeling the distribution of the background as it appears more frequently than moving objects. Optical flow is a common technique to aggregate the temporal information for vehicle detection~\cite{sun2004road} by simulating the pattern of object motion over time. Optical flow is also combined with symmetry tracking~\cite{kyo1999robust} and hand-crafted appearance features~\cite{cui2010vehicle} for better performance. However, this kind of approach is unable to distinguish the fine-grained categories of the moving objects such as car, bus, van or person. In addition, these methods need lots of complex post-processing algorithms like shadow detection and occluded vehicle recognition to refine the detection results.

Then, the statistical learning methods based on the hand-crafted features are applied to detect the vehicles from the images directly. They first describe the regions of the image by some feature descriptors and then classify the image regions into different classes such as vehicle and non-vehicle. Features like HOG~\cite{yuan2011learning, sun2006monocular}, SURF~\cite{hsieh2014symmetrical}, Gabor~\cite{sun2006monocular} and Haar-like ~\cite{chang2010online, sivaraman2010general} are commonly used for vehicle detection followed by classifiers like SVM~\cite{yuan2011learning,hsieh2014symmetrical}, artificial neural network~\cite{sun2006monocular} and Adaboost~\cite{ chang2010online, sivaraman2010general}. More advanced algorithms like DPM~\cite{niknejad2012road, yebes2014supervised} and And-Or Graph ~\cite{li2014integrating,wu2016learning} explore the underlying structures of vehicles and use hand-crafted features to describe each part of vehicles. These features, however, have limited ability of feature representation, which is difficult to handle complex scenarios.

Recently, features learned by the deep convolutional neural network show strong representation of the semantic meanings of objects, which make a great contribution to the state-of-the-art object detectors~\cite{girshick2014rich,girshick2015fast,ren2015faster,xiang2017subcategory}.
Although these methods overperform lots of hand-crafted vehicle detection methods on the vehicle detection benchmark~\cite{geiger2012we}, the vehicles with a large variance of scales (Fig.~\ref{fig:motivation} and Fig.~\ref{fig:arc}) are still difficult to be detected accurately in a real-time manner due to the scale sensitive convolutional features.
We will elaborate the scale-sensitive problem of current CNNs in the following section.

\section{Why Current CNNs are Scale-Sensitive}

It is well-known that CNNs are sensitive to scale variations in detection tasks~\cite{cai2016unified}.
In this section, we first carefully analyze its underlying reasons and then discuss how existing solutions solve this problem.

\subsection{Structure Distortion Caused by RoI Pooling} \label{subsec:structure_distortion}

CNNs-based object detection algorithms fall into two categories.
The first category is built upon a two-stage pipeline~\cite{girshick2015fast,girshick2014rich,he2016deep,kim2016pvanet,kong2016hypernet,li2016r,lin2016feature,ren2015faster,shrivastava2016training,xie2016aggregated}, where the first stage extracts proposals and the second stage predicts their classes.
The second category aims to train an end-to-end object detector~\cite{liu2016ssd,redmon2016you,redmon2016yolo9000}, which skips the object proposal detection and hence has relatively faster computational speed.
Such a detector first implicitly divides the image into a grid, then simultaneously makes prediction for each square or rectangle in the grid, and finally figures out the bounding boxes of targeting objects based on the predictions of the squares or rectangles~\cite{redmon2016you}.
However, this grid-based paradigm cannot obtain comparable accuracy to two-stage detection pipeline, as the grids have too strong spatial constraints to predict small objects appeared as groups~\cite{li2016r}.
In this regard, most of the existing methods employ the two-stage detection pipeline.

\begin{figure}[t]
	\begin{center}
		\includegraphics[width=1.0\linewidth]{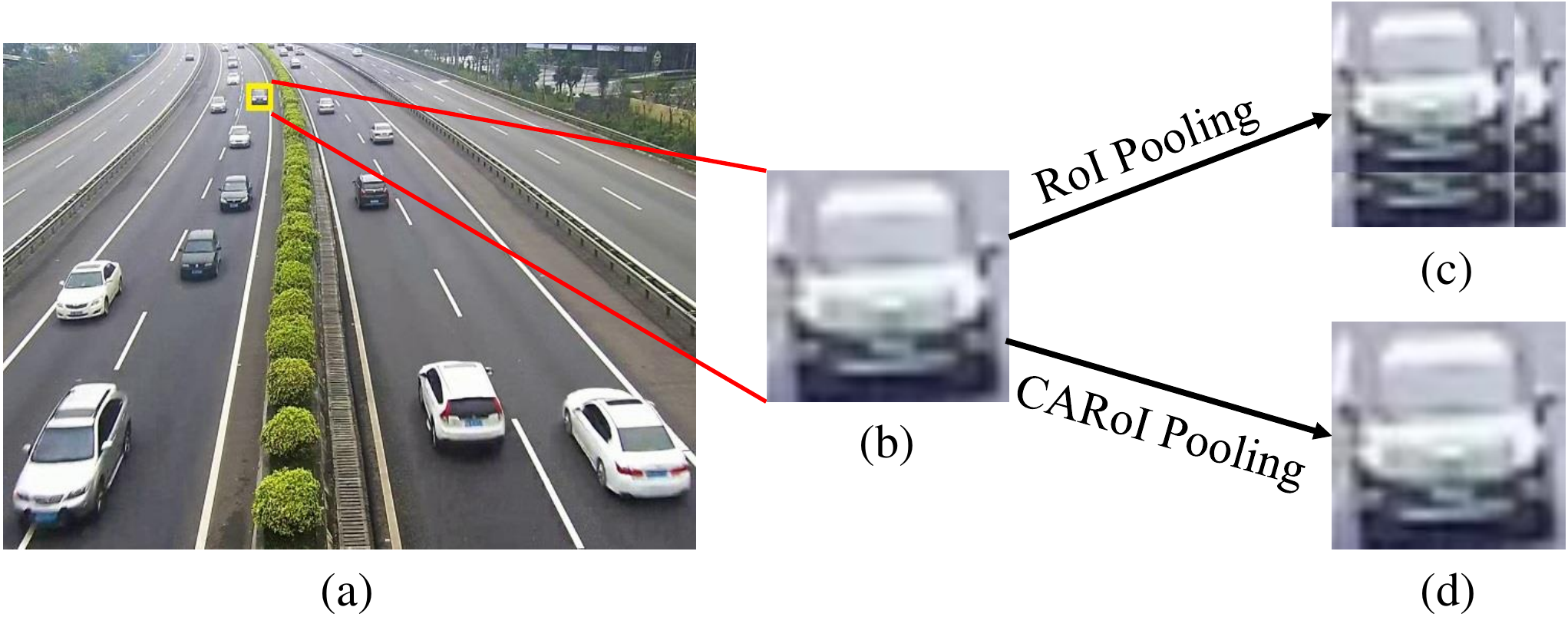}
	\end{center}
    \vspace{-5mm}
	\caption{The difference between RoI pooling and CARoI pooling. For the sake of clarity, we apply these two pooling layers on natural images instead of feature maps.}
	\label{fig:RoIPoolingSARoIPooling}
	\vspace{-5mm}
\end{figure}

In order to satisfy the input requirement of the classification networks, most two-stage object detection algorithms, e.g., SPP~\cite{he2015spatial}, Fast RCNN~\cite{girshick2015fast} and Faster RCNN~\cite{ren2015faster}, represent each proposal as a fixed-size feature vector by RoI pooling~\cite{girshick2015fast}.
As shown in Fig.~\ref{fig:motivation} (c), the RoI pooling divides every proposal into $H\times W$ sub-windows and uses the max pooling to extract one value for each sub-window so that the output can have a fixed-size of $H\times W$.
If a proposal is smaller than $H\times W$, it is enlarged to $H\times W$ by simply replicating some parts of the proposal to fill the extra space.
Unfortunately, such a scheme is not appropriate as it may destroy the original structures of the small objects (see Fig.~\ref{fig:RoIPoolingSARoIPooling} (c)).
During the network training process, filling with replicated values not only leads to inaccurate representations in the forward propagation, but also accumulates errors in the backward propagation. The inaccurate representations and accumulated errors mislead the training and prevent the network from correctly detecting small scale vehicles.
In our experiments, we find that this problem is critical for the low detection accuracy of small vehicles.

\begin{figure*} [htpb]
	\begin{center}
		\includegraphics[width=1.0\linewidth]{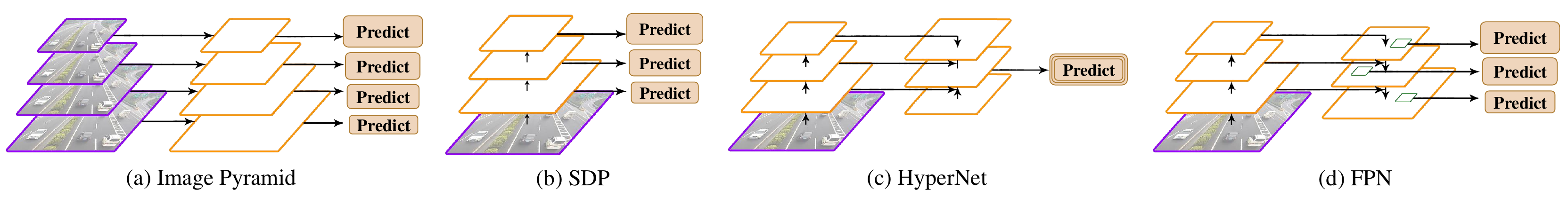}
	\end{center}
	\vspace{-6mm}\caption{(a) Multiple predictions based on multi-scale images. (b) Multiple predictions on multi-layer features.
		(c) Single prediction on concatenation features.
		(d) Multiple predictions on multi-layer features concatenating with low-layer features.
	}
	\label{fig:StrategyMultiScale}
	\vspace{-3mm}
\end{figure*}

\subsection{Intra-class Distance Caused by Scale Variations}\label{s:2.3}

The other important issue that causes scale sensitivity is the large intra-class distance between large and small scale objects.
Once the features of each proposal are extracted, they are fed into a decision network for classification.
Existing methods treat objects within the same class equally regardless of their scales.
We argue that this may lead to inaccurate detection, as the intra-class distance between large and small scale objects may be as significant as the intra-class distance on their feature representations.

\subsection{Existing Solutions and Their Shortcomings}

A lot of effort has been dedicated to solving this scale sensitivity issue.
As mentioned, most existing solutions are designed based on two types of pyramid representations.
The first one applies the concept of image pyramid (Fig.~\ref{fig:StrategyMultiScale} (a)), which exploits the multi-scale input images to make the network fit all the scales \cite{girshick2015fast,he2016deep,sermanet2013overfeat,shrivastava2016training,xiang2017subcategory,xie2016aggregated}.
However, the main disadvantage of this representation is its large computational cost~\cite{lin2016feature}, prohibiting its application to real-time detection tasks.

The other representation is the feature pyramid, which exploits the information extracted from multi-layer feature maps.
The first and straightforward attempt is to use the high resolution shallow layers to detect small objects, while using low resolution deep layers to detect large objects (as shown in Fig.~\ref{fig:StrategyMultiScale} (b)).
This strategy has been adopted by SSD~\cite{liu2016ssd}, MSCNN~\cite{cai2016unified}, FCN~\cite{shelhamer2017fully} and SDP~\cite{yang2016exploit}.
However, as the feature maps in the shallow layers lack of the semantic information, they usually fail to distinguish the small objects accurately.

In order to take full advantage of deep layer information to tackle scale variations, some researchers presented to combine multi-layer feature maps together to train a network (e.g., MultiPath~\cite{zagoruyko2016multipath} and HyperNet~\cite{kong2016hypernet}, see Fig.~\ref{fig:StrategyMultiScale} (c)).
However, due to the down-sampling operations used in the network, small objects cannot maintain sufficient spatial information in the deep layers, and thus they are still difficult to be detected.
To better maintain the deep feature maps of small objects, another solution is proposed to use the high-resolution feature maps and the up-sampled deep feature maps together to predict small objects, such as~\cite{lin2016feature,shrivastava2016beyond} (Fig.~\ref{fig:StrategyMultiScale} (d)).
The main problem of this solution is that the upsampling operation is performed on the entire feature map, which requires more memory resources and additional computational costs.
While extra information leads to better accuracy, high computational cost is inevitable~\cite{huang2016speed}, which is unacceptable for our real-time vehicle detection task.

As a consequence, instead of introducing additional steps to solve the scale-sensitive problem, we aim to address this problem internally by introducing two simple solutions: a novel context-aware pooling and a multi-branch decision network, which lead to zero extra computational cost while effectively dealing with the scale sensitivity issue for real-time and accurate vehicle detection.

\section{Scale-Insensitive Network}
\subsection{Overview}

The architecture of the proposed scale-insensitive network (SINet) is illustrated in Fig.~\ref{fig:arc}.
Our SINet takes the whole image as input and outputs the detection result in an end-to-end manner.
It first generates a set of convolutional feature maps~\cite{lecun2015deep} and obtains a set of proposals based on these feature maps using region proposal networks (RPN)~\cite{cai2016unified,ren2015faster}.
The RPN predicts the bounding boxes that have a large probability of containing objects and these predicted bounding boxes are called as proposals.
Then, the proposed context-aware RoI pooling (CARoI pooling) is used to extract the features for each proposal. 
The CARoI pooling applies the deconvolution with bilinear kernels to enlarge the feature regions of the small proposals to avoid representing the small objects with the replicated values. 
The CARoI pooling is applied to multiple layers of the CNN and these pooled features at different layers are concatenated together to fuse the low-level detail information and the high-level semantic information to detect the objects~\cite{kong2016hypernet}.
After that, we split the SINet into multiple branches according to the sizes of the proposals, alleviating the training burden for the large intra-class variation of objects with different scales.
In this case, we can improve the detection precision for both large objects and small objects.
Lastly, we fuse all the predicted results from multiple branches into the final detection result.
The deconvolution with bilinear kernels and the multi-branch decision network do not increase processing time because the former just deals with small proposals without enlarging the whole feature maps, and the latter processes the same number of proposals as traditional detection methods.

\subsection{Context-aware RoI Pooling}
\label{sec:CARoIpooling}

The context-aware RoI pooling (CARoI pooling) can 
adjust the proposals to the specified size without sacrificing important contextual information (as illustrated in Fig.~\ref{fig:RoIPoolingSARoIPooling} (d)).

In CARoI pooling, we have three cases to deal with. Firstly, if the size of a proposal is larger than the specified size, we shall extract the maximum value in each sub-window as original RoI pooling strategy (as described in Section~\ref{subsec:structure_distortion}). Secondly, if the size of a proposal is smaller than the specified size, a deconvolution operation with bilinear kernel is applied to enlarge the proposal while keeping the circumstances from being impaired so that we can still extract discriminative features from the small proposals.
The size of deconvolution kernel is dynamically determined by the proposal size and the predefined pooled size. Specifically, the kernel size is equal to the ratio between the specified size of pooled feature map and the size of each proposal.
Thirdly, when the width of a proposal is larger than the pooled length and the height of this proposal is smaller than the pooled length
, our CARoI pooling applies the deconvolution operation to enlarge the height of this proposal, splits the width of this proposal into several sub-windows (the number of the sub-windows is equal to the pooled length) and uses the maximum value of each sub-window as the most discriminative feature value.

Mathematically, we formulate the three cases mentioned above in the following equations.
Let $y_{k}^{j}$ be the $j$-th output of CARoI pooling layer from the $k$-th proposal.
The CARoI pooling computes $y_{k}^{j}=x_{i^*}$, where:
\begin{equation} \label{eq3}
i^*=argmax_{i\in R(k,j)}x_{i}
\end{equation}
\begin{equation}
x_{i}\in(X_{k}\otimes\sigma_{k})
\end{equation}
In above equations, $R(k,j)$ represents the index set of the sub-window where the output unit $y_{k}^{j}$ selects the maximum feature value. $x_i\in\mathbb{R}$ is the $i$-th feature value on the feature map. And we use the $X_{k}$ to represent a set of input features of $k$-th proposal. $\otimes$ denotes the deconvolution operation and $\sigma_{k}$ is the kernel of the deconvolution operation, which is determined by the scales of proposals. If the size of proposal is less than the pooled feature map size, this deconvolution kernel is equal to the ratio between the specified size of pooled feature map and the size of each proposal; otherwise, this deconvolution kernel is equal to one, which suggests this deconvolution operation doesn't take effect on the large proposals.
After obtaining the discriminate features, the maximum values of these features in each sub-window are used to represent this proposal.

\textbf{Back-propagation.} Derivatives are diverted through CARoI pooling by back propagation to train the network. The partial derivative of loss $L$ respective to input variable $x_{i}$ is:
\begin{equation}
\frac{\partial L}{\partial x_{i}}=\sum_{k}\sum_{j}[i=i^*]\nabla_{\sigma_{k}}(\frac{\partial L}{\partial y_{k}^{j}})
\end{equation}
where $i^*$ is the index described in Equation~\ref{eq3}, which indicates the position of maximum values in each sub-window after the deconvolution.
$\nabla_{\sigma_{k}}(\frac{\partial L}{\partial y_{k}^{j}})$ indicates the derivative of the deconvolution with respect to the loss  $\frac{\partial L}{\partial y_{k}^{j}}$. This loss is propagated from following layers that are connected to CARoI pooling. This derivative will be accumulated by all RoIs and all positions ($\sum_{k}\sum_{j}$).

\subsection{Multi-branch Decision Network} \label{sb:3.3}

As analyzed in Section~\ref{s:2.3}, another critical issue for CNN-based object detection is the large scale variations of targeting object,
which is common in vehicle detection.
To reduce the scale variance of objects,
we present to split the proposals with different sizes into different branches and each branch is used to detect a set of objects with similar sizes.
Each branch consists of one convolutional layer and one fully connected layer followed by two classifiers: one is used for classification; the other is used for bounding box regression (see \cite{girshick2015fast} for details).
Although we split the proposals into multiple branches, these proposals share the features extracted by some convolutional layers (as the blue boxes shown in Fig.~\ref{fig:arc}).

The number of branches is empirically determined by considering the scale distribution of the dataset and the computational resources, which is discussed in section~\ref{4.4}. Here we take the two-branch decision network as an example but this technique can be easily extended to multi-branch decision network. In two-branch decision network, we mainly use the median value of all objects' scales in the training set as the reference threshold to split the proposals into the large branch or the small branch. During the training process, in order to make two branches share a portion of samples in the median scales and augment the size of training samples for each branch, the threshold for splitting proposals is dynamically changed in each training iteration. We simulate the threshold change by a Gaussian model, and the median value of all objects' scales is the mean value of the Gaussian model. 
In such a way, those proposals with the scales that are near to the median value of all objects' scales have opportunity to be categorized into the large and the small branches in the whole training procedure.
In testing, we simply use median value to split the proposals.

\subsection{Implementation Details}

\textbf{Network architecture.}
In principle, our context aware RoI pooling and multi-branch decision network are general and can be built on any CNN architectures.
In this paper, we test our algorithms based on the PVA network~\cite{kim2016pvanet} and VGG network~\cite{simonyan2014very}.
The kernel sizes of CARoI pooling are set to $6\times6$ in the PVA network and $7\times7$ in the VGG network.

We use the proposal extraction network (RPN) proposed by MS-CNN~\cite{cai2016unified} to extract high-quality proposals from different layers of the CNN. Then we connect the multi-branch decision network at the end of the RPN to build the SINet as shown in Fig.~\ref{fig:arc}.
The whole network is trained in an end-to-end manner.

\textbf{Training strategies.}
Stochastic gradient descent (SGD) is used to optimize our SINet.
In order to make the training process stable, we first harness a small learning rate to train the RPN and then leverage a large learning rate to train the whole network end to end.
We first set the learning rate as 0.0001 for 10k iterations with weight decay 0.0005 to train the RPN. Then, to train the whole network, we set the learning rate as 0.0005, reduce it by a factor of 0.1 at 40k and 70k iterations and stop learning after 75k iterations.
If employing VGG net, we adjust the initial learning rate to 0.00005 and 0.0001 for the first and second stages respectively.
To accelerate training and reduce overfitting~\cite{yosinski2014transferable}, the weights of convolutional layers in VGG trained on ImageNet~\cite{russakovsky2015imagenet} or PVA trained on Pascal VOC~\cite{everingham2010pascal} are used to initialize the RPN. Then, we utilize the well-trained weights of the fully connected layers in VGG and PVA to initialize the fully connected layers of the newly added multi-branch decision network. Other layers are initialized by random noise.
There are four images in each batch.
In addition, 
data argumentation methods and hard example mining strategies~\cite{cai2016unified} are also used as in the MS-CNN.

\textbf{Inference.}
In testing, for each input image, the network produces outputs of small and large objects in multiple branches.
Then we combine them together and use non-maximum suppression (NMS) to refine the results.
Instead of selecting only the bounding box with the maximum confidence from highly overlapping detection boxes, we choose several bounding boxes with the relatively high confidences among these boxes and average the coordinates of them.
We call this strategy as \emph{soft-NMS}, which is useful to improve the localization accuracy for occluded vehicles.

\section{Experiments}

In order to evaluate the effectiveness of the proposed SINet, we conduct experiments on two representative vehicle datasets: the KITTI dataset and a newly constructed large scale variance highway dataset (LSVH).
The experiments are implemented on Ubuntu 14.04 with a single GPU (NVIDIA TITAN X) and 8 CPUs (Inter(R) Xeon(R) E5-1620 v3 @ 3.50GHz).

\subsection{Datasets and Evaluation Metrics} \label{4.1}

\textbf{KITTI dataset.} KITTI~\cite{geiger2012we} is a widely used benchmark for vehicle detection algorithms. It contains various scales of vehicles in different scenes. It consists of $7481$ images for training and $7518$ images for testing.
According to size, occlusion and truncation, the organizers classify these targeting vehicles into three difficulty levels: easy, moderate and hard; check~\cite{geiger2012we} for detailed definition of these difficulty levels.

\begin{figure}[tp]
	\begin{center}
		\includegraphics[width=1.0\linewidth]{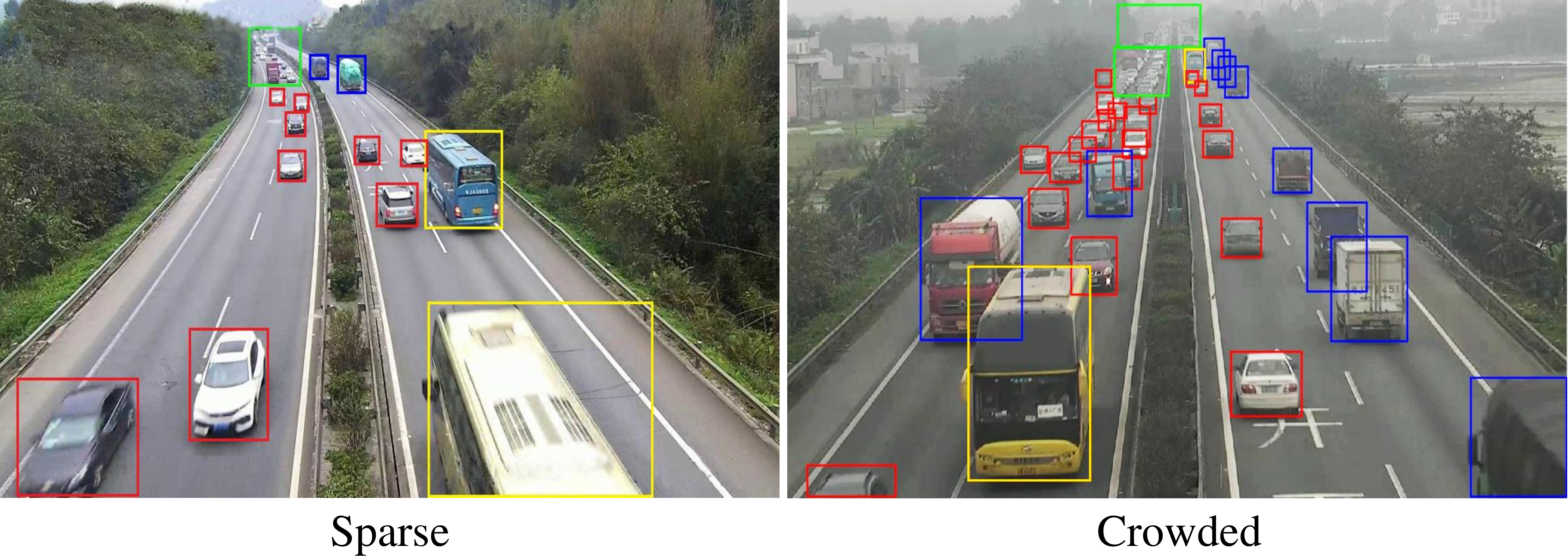}
	\end{center}
	\vspace{-5mm}\caption{Examples of our large scale variance highway (LSVH) dataset under the different scenes. Red, blue, orange and green boxes are labelled to indicate the ``car'', ``van'', ``bus'', and ``don't care'' regions respectively. }
	\label{fig:highway-dataset}
	\vspace{-2mm}
\end{figure}

\begin{table} [tp]
	\renewcommand{\arraystretch}{1.3}
	\caption{Data distribution on LSVH dataset.}
	\label{table:LSVH}
	\centering
	\begin{tabular}{c|c|c|c}
		\hline
		& Sparse & Crowded &	Total   \\
		
		\hline
		Image & 12979 & 1409 & 14388 \\ 
		\hline
		Car & 40025 & 35116 & 75141 \\ 
		Bus & 4419 & 2480 & 6899 \\  
		Van & 10474 & 4812 & 15286\\ 
		
		\hline
		Vehicle/Image & 4.23  & 30.10 & 6.76 \\
		\hline
		
	\end{tabular}
	\vspace{-4mm}
\end{table}

\textbf{LSVH dataset.}
Highway is a typical road scene that contains vehicles with large scale variations, as the surveillance cameras usually cover a large and long view of the road.
We construct a new large-scale variance highway dataset, which contains 16 videos captured under different scenes, time, weathers and resolutions, as shown in Fig. \ref{fig:highway-dataset} and~\ref{fig:Visulization}.
As illustrated in Fig.~\ref{fig:highway-dataset} and Table~\ref{table:LSVH}, the vehicle is classified into three categories (car, bus and van) under two scenes (sparse and crowded).
We consider a video scene as a crowded scene in case that it contains more than $15$ vehicles per frame on average; otherwise, it is considered as a sparse scene.
For the vehicles that are too small to be recognized by human are labelled as ``don't care'', and these regions are ignored during training and evaluation. Specifically, an object whose height is less than 15 pixels will be ignored. There are 75141 cars, 6899 buses and 15286 vans in total in our LSVH.
We use two strategies to split the videos into training/testing sets: (1): we separate each video into two parts, and select the first seventy percentages of each video as the training data and the remaining thirty percentages as the testing data; 
(2) we use the eight videos in ``Sparse'' as the training data, and the left four videos in ``Sparse'' as the testing data.
To avoid retrieving similar images, we extract one frame in every seven frames of these videos as the training/testing images.

\textbf{Evaluation metrics.}
We employ the well established average precision (AP) and intersection over union (IoU) metrics ~\cite{everingham2010pascal} to evaluate the performance; it has been widely used to assess various vehicle detection algorithms~\cite{everingham2010pascal,geiger2012we}.
For KITTI, we evaluate our method in all three difficulty levels.
For LSVH, we evaluate the performance for car, bus and van under the scenes of sparse and crowded, respectively. The IoU is set to 0.7 in these two datasets, which means only the overlap between the detection bounding box and the ground truth bounding box greater than or equal to 70\% is considered as a correct detection.

\subsection{Comparison with the State-of-the-arts} \label{4.2}

\begin{table} [tp]
	\renewcommand{\arraystretch}{1.3}
	\caption{Results on the KITTI benchmark. All methods are ranked based on the ``Moderate''.} 
	\label{table:KITTIBenchmark}
	\centering
	\begin{tabular}{c|c|c|c|c}
		\hline
		\multirow{2}{*}{Model} &
		\multirow{2}{*}{Time/Image}  &
		\multicolumn{3}{c}{Average Precision (\%)}   \\
		\cline{3-5}  & & Moderate & Easy & Hard   \\
		\hline
		\textbf{SINet\_VGG}  \textbf{(ours)}&	0.2s&	\textbf{89.60}&	90.60&	77.75\\
		\textbf{SINet\_PVA} \textbf{(ours)}& \textbf{0.11s} &	89.21&	91.91&	76.33 \\
		
		Deep3DBox~\cite{mousavian20163d}&	1.5s&	89.04&	92.98&	77.17 \\
		
		SubCNN~\cite{xiang2017subcategory}&	2s&	89.04&	90.81&	79.27 \\
		MS-CNN~\cite{cai2016unified}&	0.4s&	89.02&	90.03&	76.11 \\
		SDP+RPN~\cite{ren2015faster,yang2016exploit}&	0.4s&	88.85&	90.14&	78.38 \\
		Mono3D~\cite{chen2016monocular}&	4.2s&	88.66&	92.33&	78.96 \\
		3DOP~\cite{chen20153d}&	3s&	88.64&	\textbf{93.04}&	79.10 \\
		
		MV3D~\cite{chen2016multi}&	0.45s&	87.67&	89.11&	\textbf{79.54} \\
		
		SDP+CRF (ft)~\cite{yang2016exploit}&	0.6s&	83.53&	90.33&	71.13 \\
		Faster R-CNN~\cite{ren2015faster}&	2s&	81.84&	86.71&	71.12 \\
		
		MV3D (LIDAR)~\cite{chen2016multi}&	0.3s&	79.24&	87.00&	78.16 \\
		
		spLBP~\cite{hu2016fast}&	1.5s&	77.39&	87.18&	60.59 \\
		Reinspect~\cite{stewart2016end}&	2s&	76.65&	88.13&	66.23 \\
		Regionlets~\cite{long2014accurate,wang2015regionlets,zou2014generic}&	1s&	76.45&	84.75&	59.70 \\
		AOG~\cite{li2014integrating,wu2016learning}&	3s&	75.94&	84.80&	60.70 \\
		3DVP~\cite{xiang2015data}&	40s&	75.77&	87.46&	65.38 \\
		SubCat~\cite{ohn2015learning}&	0.7s&	75.46&	84.14&	59.71 \\
		\hline
		YOLOv2~\cite{redmon2016yolo9000} & \textbf{0.03s} & 61.31 & 76.79 & 50.25\\
		YOLO~\cite{redmon2016you} & \textbf{0.03s}& 35.74 &  47.69 & 29.65 \\
		\hline

	\end{tabular}
    \vspace{-4mm}
\end{table}

\begin{table*} [tp]
	\renewcommand{\arraystretch}{1.3}
	\caption{Comparison on our LSVH dataset. We use two strategies to split the dataset (see Section~\ref{4.1} for details).}
	\label{table:highway}
	\centering
	\begin{tabular}{c||c||c|c|c|c|c|c|c||c|c|c|c}
		\hline
		\multirow{3}{*}{Model} &
		\multirow{3}{*}{Time/Image} & \multicolumn{7}{c||}{Strategy 1} & \multicolumn{4}{c}{Strategy 2}  \\
	    \cline{3-13}& & \multirow{2}{*}{Mean} &
		\multicolumn{3}{c|}{Sparse}  & \multicolumn{3}{c||}{Crowded} &  \multicolumn{4}{c}{Sparse} \\
		\cline{4-13}  & & & Car & Bus & Van & Car & Bus & Van  & Mean & Car & Bus & Van \\
		\hline
		
		\textbf{SINet\_VGG} \textbf{(ours)}& 0.20s & \textbf{70.17} & \textbf{81.82} & \textbf{85.60} & \textbf{78.65} & \textbf{56.80} & \textbf{55.78} & 62.38 
		&\textbf{78.71} & 74.51 & \textbf{84.34} & \textbf{77.27}\\
		\textbf{SINet\_PVA} \textbf{(ours)}& \textbf{0.08s} & 70.04 & 81.40 & 84.39 & 77.39  & 53.76 & 54.06 & \textbf{69.23} & 77.69 & \textbf{74.79} & 81.35 & 76.92 \\
		MS-CNN~\cite{cai2016unified} & 0.23s & 63.23 & 79.94 & 83.71 & 76.79 & 51.74 & 32.95 & 54.26 & 72.66 &71.13 &74.34 & 72.50	\\
		Faster RCNN~\cite{ren2015faster} & 0.31s & 46.44 & 60.93 & 66.68 & 60.14 & 26.08 & 24.55 & 40.24 & 40.22 & 36.19 & 42.79 & 41.69  \\
		\hline
		YOLOv2~\cite{redmon2016yolo9000}& \textbf{0.03s}& 43.82 & 59.71 &65.51&58.35 & 17.39 & 21.55& 40.42 & 54.00 & 53.16 & 53.88  & 54.96 \\
		YOLO~\cite{redmon2016you} & \textbf{0.03s} & 16.53 & 23.06 & 31.13& 22.44 & 3.87 & 8.35 &  10.32 & 23.78 & 22.97 & 24.52 & 23.85\\
		\hline
		
	\end{tabular}
\vspace{-4mm}
\end{table*}

We compare the proposed SINet to the state-of-the-arts on both the KITTI dataset and our LSVH dataset. Table~\ref{table:KITTIBenchmark} shows the performance published on the KITTI website.
In this experiment, the entire training set is used for training our models, and the tested results are uploaded to the KITTI website. We compare our models with other 18 published methods. 
It is clear that our SINet achieves the highest accuracy on moderate case and fastest speed (except the one-stage deep learning based detectors, e.g. YOLO and YOLOv2, which are fast but with very low accuracy). Our method can achieve the same speed as YOLO and YOLOv2 by reducing the size of input images and the accuracy is still much better than these two methods (see Section~\ref{4.3}).
For the computational efficiency among the two-stage deep learning based detectors, our SINet takes only $1/14$ of Deep3DBox~\cite{mousavian20163d}.

Table~\ref{table:highway} shows the performance on our LSVH dataset.
It is obvious that both two variants of our SINet outperforms the MS-CNN baseline and Faster RCNN in terms of detection accuracy and efficiency.
Our SINet also surpasses the one-stage detectors (YOLO and YOLOv2) by a significant margin for the accuracy.
In particular, the SINet shows a good performance to detect the vehicles under the ``Crowded'' scene.

\begin{figure}[tp]
	\begin{center}
		\includegraphics[width=1.0\linewidth]{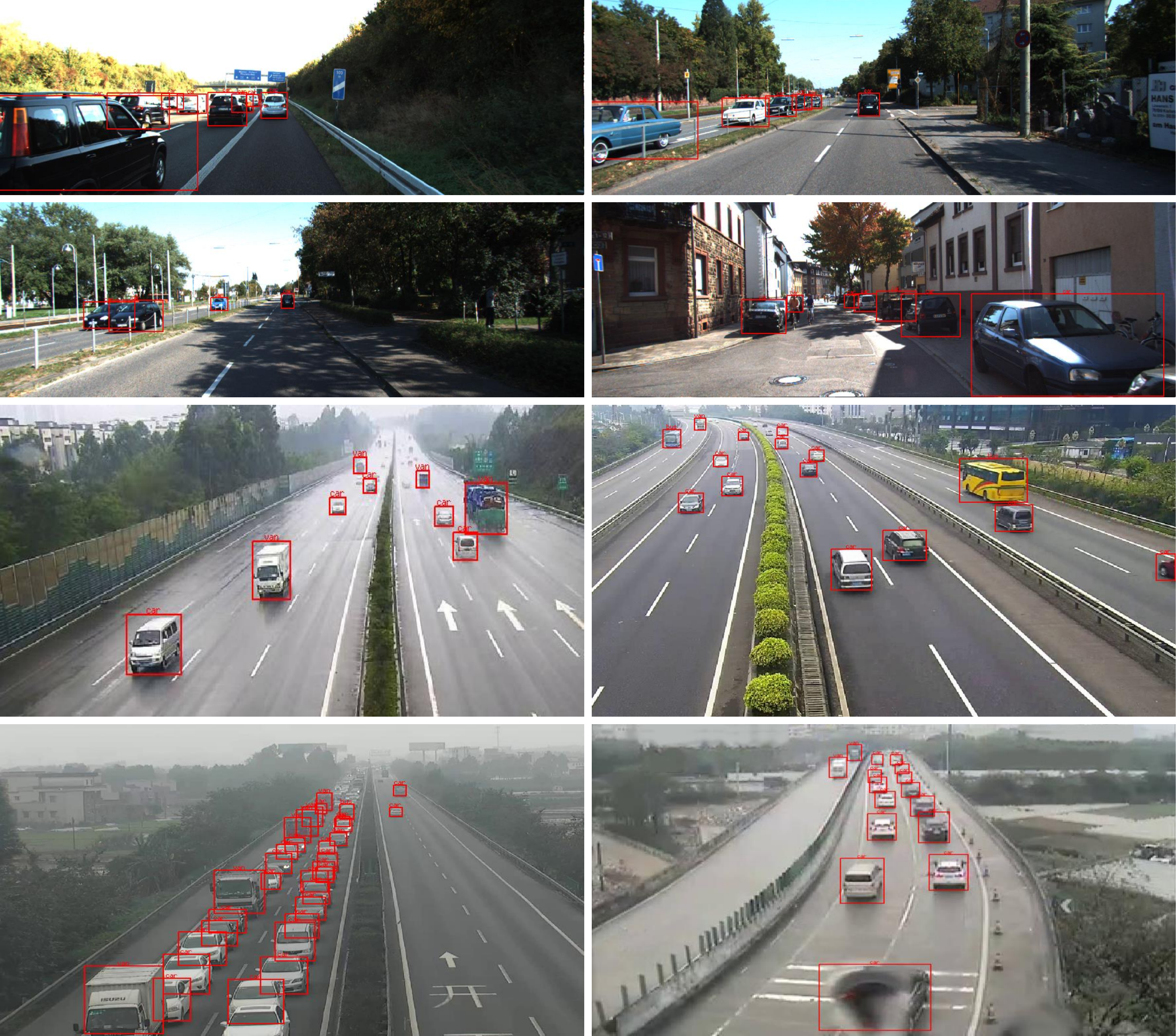}
	\end{center}
	\vspace{-2mm}\caption{Examples of detection results by our SINet on the KITTI dataset (the first two rows) and our LSVH dataset (the last two rows). (Best viewed in color and at full size on a high-resolution display.)}
	\label{fig:Visulization}
	\vspace{-5mm}
\end{figure}

In Fig.~\ref{fig:Visulization}, we visualize the vehicles detected by SINet on the images from KITTI dataset and our LSVH dataset. It is clear that our algorithm is effective to detect the vehicles with different orientations, scales, truncation levels under the different situations such as blurry, rainy, and occluded. Moreover, our SINet shows a strong ability on detecting vehicles with a large variance of scales, especially for small vehicles.
This corroborates that the presented SINet has potential to be used as a powerful tool for intelligent transportation systems.

\subsection{Image Resolution Sensitivity} \label{4.3}

\begin{figure}[tp]
	\begin{center}
		\includegraphics[width=0.8\linewidth]{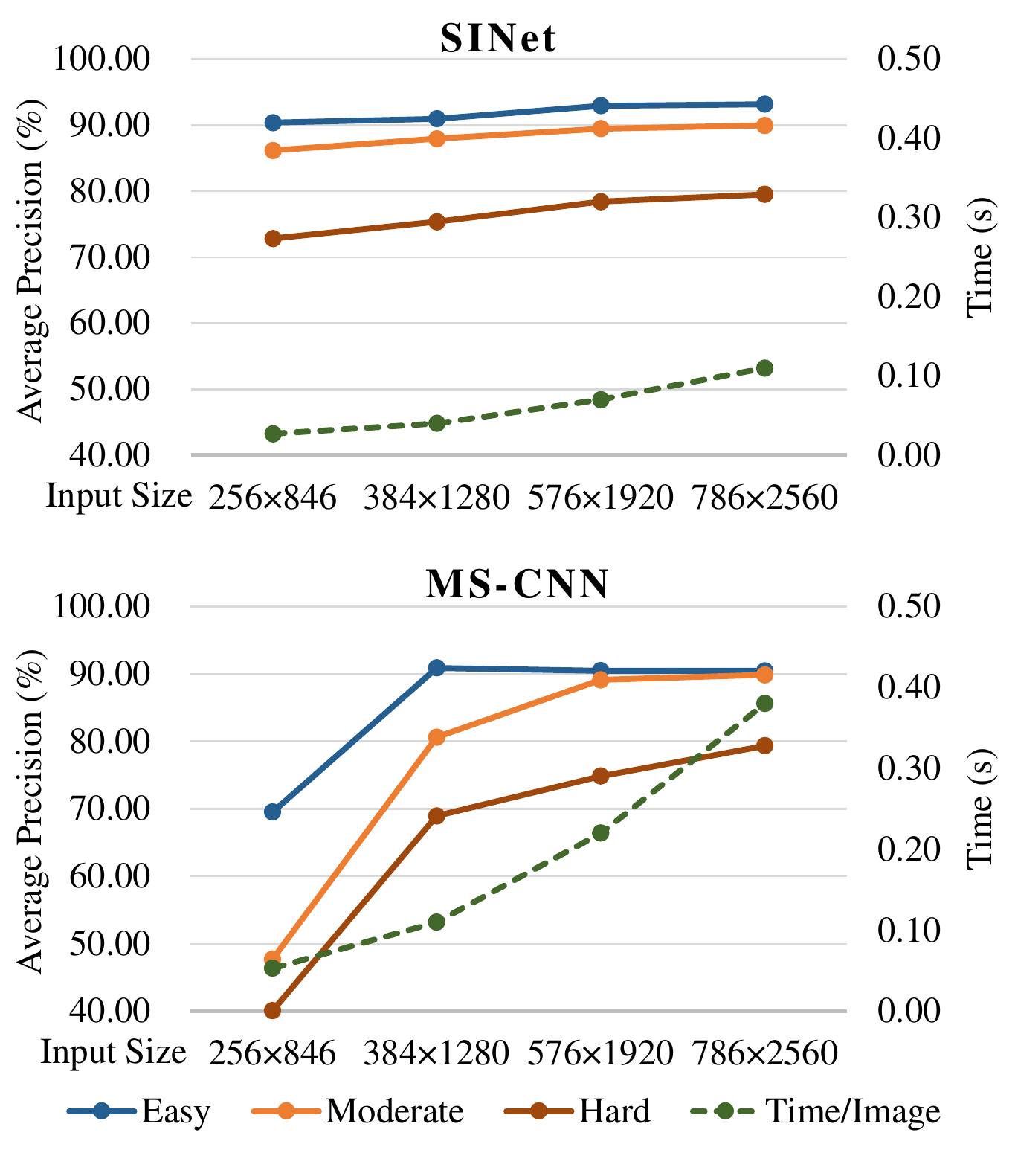}
	\end{center}
	\vspace{-5mm}\caption{Image resolution sensitivity evaluation. These experiments were done on the KITTI training set.}
	\label{fig:InputSize}
	\vspace{-5mm}
\end{figure}

Since our SINet has a strong capability on feature representation for low resolution vehicles, it can also perform well on the low resolution images. This resolution insensitive property is actually very important for practical usage, and enabling fast computation by resizing an image to a small resolution. Fig.~\ref{fig:InputSize} illustrates our SINet is insensitive to the image resolution. The detection performance is robust with different sizes of input images. On the contrary, MS-CNN~\cite{cai2016unified} is sensitive to the resolution of input images. A small input image decreases the accuracy dramatically, while increasing the input resolution leads to much more computational overhead.

\subsection{Ablation Analysis} \label{4.4}

\begin{table*} [htbp]
	\renewcommand{\arraystretch}{1.3}
	\caption{Ablation analysis of the presented SINet on the KITTI validation set. The time is evaluated on a single NVIDIA TITAN X GPU (maxwell version) with $12GB$ memory and only one image is processed at the same time.}
	\label{table:KITTIValidation}
	\centering
	\begin{tabular}{c|c|c|c|c|c|c|c}
		\hline
		\multirow{2}{*}{Model} &
		\multirow{2}{*}{Network} &
		\multirow{2}{*}{Number of Branches} &
		\multirow{2}{*}{Post Processing} & \multicolumn{3}{c|}{Average Precision (\%)}  &	\multirow{2}{*}{Time/Image} \\
		\cline{5-7} & & & & Moderate & Easy & Hard & \\
		\hline
		\hline
		MS-CNN\_PVA & PVA & 1 & NMS & 82.69 & 91.85 & 69.74 &  \textbf{0.07s} \\
		+CARoI pooling & PVA & 1 & NMS & 88.39 & 92.52 & 75.70 & \textbf{0.07s} \\
		+Multi-branch decision network & PVA & 2 & NMS & 89.36 & 92.96 & 78.11 &  \textbf{0.07s} \\
		\textbf{SINet\_PVA} & PVA & 2 & Soft-NMS & 89.49 & 92.95 & 78.45 & \textbf{0.07s} \\
		
		\textbf{SINet\_PVA} & PVA & 3 & Soft-NMS & \textbf{89.53} & \textbf{93.31} & \textbf{78.53} & \textbf{0.07s} \\
		\hline
		MS-CNN\_VGG~\cite{cai2016unified} & VGG & 1 & NMS & 89.13 & 90.49 & 74.85  & 0.22s \\
		+CARoI pooling & VGG & 1 & NMS & 90.07 & 95.30 & 79.31 & \textbf{0.20s} \\
		+Multi-branch decision network & VGG & 2 & NMS & 90.22 &  95.82 & 80.02 & \textbf{0.20s} \\
		\textbf{SINet\_VGG} & VGG & 2 & Soft-NMS & \textbf{90.33} &  \textbf{95.84} & \textbf{80.14} & \textbf{0.20s} \\
		\textbf{SINet\_VGG} & VGG & 3 & Soft-NMS & - & - & - & Out of memory \\
		\hline
		
	\end{tabular}
    \vspace{-2mm}
\end{table*}

\begin{table*} [htbp]
	\renewcommand{\arraystretch}{1.3}
	\caption{Vehicle scale analysis on LSVH testing set (strategy 1, sunny). The size of the input image is $768\times 1344$.}
	\label{table:VehicleScale}
	\centering
	\begin{tabular}{c|c||c|c|c|c||c|c|c|c}
		\hline
		
		\multicolumn{2}{c||}{Scene} & \multicolumn{4}{c||}{Sparse} & \multicolumn{4}{c}{Crowded} \\
		\hline
		\multicolumn{2}{c||}{Model} & SINet\_PVA & MSCNN\_PVA & SINet\_VGG & MSCNN\_VGG &
		SINet\_PVA & MSCNN\_PVA & SINet\_VGG & MSCNN\_VGG \\
		\hline
		\hline
		
		\multirow{3}{*}{Small} &Car &70.25&47.39&\textbf{72.46}&68.91& 13.14&2.38 &\textbf{14.50}&9.00\\
		&Bus &55.06&30.22&\textbf{59.54}&57.37& 20.10& 3.48&\textbf{22.36}&5.37\\
		&Van &48.34&25.84&\textbf{51.40}&48.78& -& -&-&-\\
		\hline
		
		\multirow{3}{*}{Medium}
		&Car &87.68&74.45&\textbf{88.44}&85.54& 55.60&22.46 &\textbf{61.06}&55.98 \\
		&Bus &86.02&70.46&\textbf{90.31}&86.13&25.45& 10.48&\textbf{29.04}& 12.18\\
		&Van &73.63&57.67&\textbf{76.08}&75.20& \textbf{17.16}& 3.87&10.16&3.58 \\
		\hline
		
		\multirow{3}{*}{Large}
		&Car &\textbf{84.72}&78.94&83.13&77.27&82.26 &59.18 &\textbf{83.25}& 80.51\\
		&Bus &90.13&82.01&\textbf{90.93}&88.75& \textbf{65.34}& 44.94&60.71& 40.43\\
		&Van &81.96&74.22&\textbf{85.16}&79.22& \textbf{77.20}& 66.86& 70.04& 63.51\\
		\hline
		
	\end{tabular}
    \vspace{-5mm}
\end{table*}

We perform ablation analysis of SINet on the KITTI dataset to evaluate how different components affect the detection performance.
As there is no ground truth provided for the testing set of KITTI, we follow~\cite{cai2016unified} to split the training set into training and validation sets, and all of them are resized to $576\times 1920$. 

Table~\ref{table:KITTIValidation} shows the experimental results.
First, comparing with the baselines which are constructed by the MS-CNN framework with PVA (the $1^{st}$ row) or VGG (the $6^{th}$ row) network, the CARoI pooling dramatically improves the accuracy while no extra time is introduced, as shown in the $2^{nd}$ and $7^{th}$ row.
Particularly, the improvements on ``Moderate'' and ``Hard'' categories are significant, which implies the recovered high resolution semantic features are very useful for detecting small objects.
Moreover, our multi-branch decision network with two branches further enhances the accuracy while keeping the efficiency, as shown in the $3^{rd}$ and $8^{th}$ row.
The soft-NMS post-processing contributes to the ``Hard'' category (the $4^{th}$ and $9^{th}$ row), which includes many occluded and truncated vehicles, demonstrating its effectiveness for occlusion and truncation cases.
When we continue to increase the number of branches, the performance gain is limited but with more network parameters which occupy more memory. 
In this case, two-branch decision network is used in other experiments.

\subsection{Vehicle Scale Analysis.} \label{4.5}
We explore the detection performance of SINet on different scales of vehicles.
This experiment is performed on our LSVH dataset which contain vehicles with a large variance of scales (as illustrated in Fig.~\ref{fig:highway-dataset}).
All vehicles are divided into three categories: ``Small'', ``Medium'' and ``Large'' based on the their scales. Specifically, the vehicles whose heights are greater than $15$ pixels and smaller than $39$ pixels belong to the ``Small'' category; the vehicles with the height between $39$ pixels and $66$ pixels are in ``Medium'' category; other vehicles with height greater than $66$ pixels belong to ``Large'' category.

As shown in Table~\ref{table:VehicleScale}, our SINet shows improvements on all scales of vehicles under the different scenes based on both PVA network and VGG network.
The improvement on small vehicles is more significant compared with other sizes of vehicles, since the baseline methods introduce more artifacts and distortions (caused by the traditional RoI pooling) to small vehicles, which can be avoided by the CARoI pooling.
Moreover, our SINet also achieves a dramatic improvement on the crowded scenes, especially for the vehicle with a small or medium scale. It shows that our approach is effective even under the complex situation, as shown in Fig.~\ref{fig:Visulization}. However, the detection accuracy for the small scale crowded vehicles is still not satisfactory. This is because these objects are highly occluded, blurry and extremely small (Fig.~\ref{fig:highway-dataset} and Fig.~\ref{fig:Visulization}).

\section{Conclusion}

In this paper, we present a scale-insensitive network, denoted as SINet, for fast detecting vehicles with a large variance of scales. Two new techniques, context-aware RoI pooling and multi-branch decision network, are presented to maintain the original structures of small objects and minimize the intra-class distances among objects with a large variance of scales.
Both of the techniques require zero extra computational effort.
Furthermore, we construct a new highway dataset which contains vehicles with large scale variance.
To our knowledge, it is the first large scale dataset focuses on the highway scene.
Our SINet achieves state-of-the-art performance on both accuracy and speed on KITTI benchmark and our LSVH dataset.
Further investigations include evaluating the SINet on more challenging datasets and integrating it into some intelligent transportation systems.

\section*{Acknowledgments}
The work was supported by NSFC (Grant No. 61772206, U1611461, 61472145, 61702194), Special Fund of Science and Technology Research and Development on Application from Guangdong Province (Grant No. 2016B010124011, 2016B010127003), Guangdong High-level Personnel of Special Support Program (Grant No. 2016TQ03X319), the Guangdong Natural Science Foundation (Grant No. 2017A030311027, 2017A030312008), the Major Project in Industrial Technology in Guangzhou (Grant No. 2018-0601-ZB-0271), 
and the Hong Kong Polytechnic University (Project no. 1-ZE8J). 
Xiaowei Hu is funded by the Hong Kong Ph.D. Fellowship.

{\small
  \bibliographystyle{IEEEtran}
  \bibliography{reference}
}

\vspace{-5mm}

\begin{IEEEbiography}[{\includegraphics[width=1in,height=1.25in,clip,keepaspectratio]{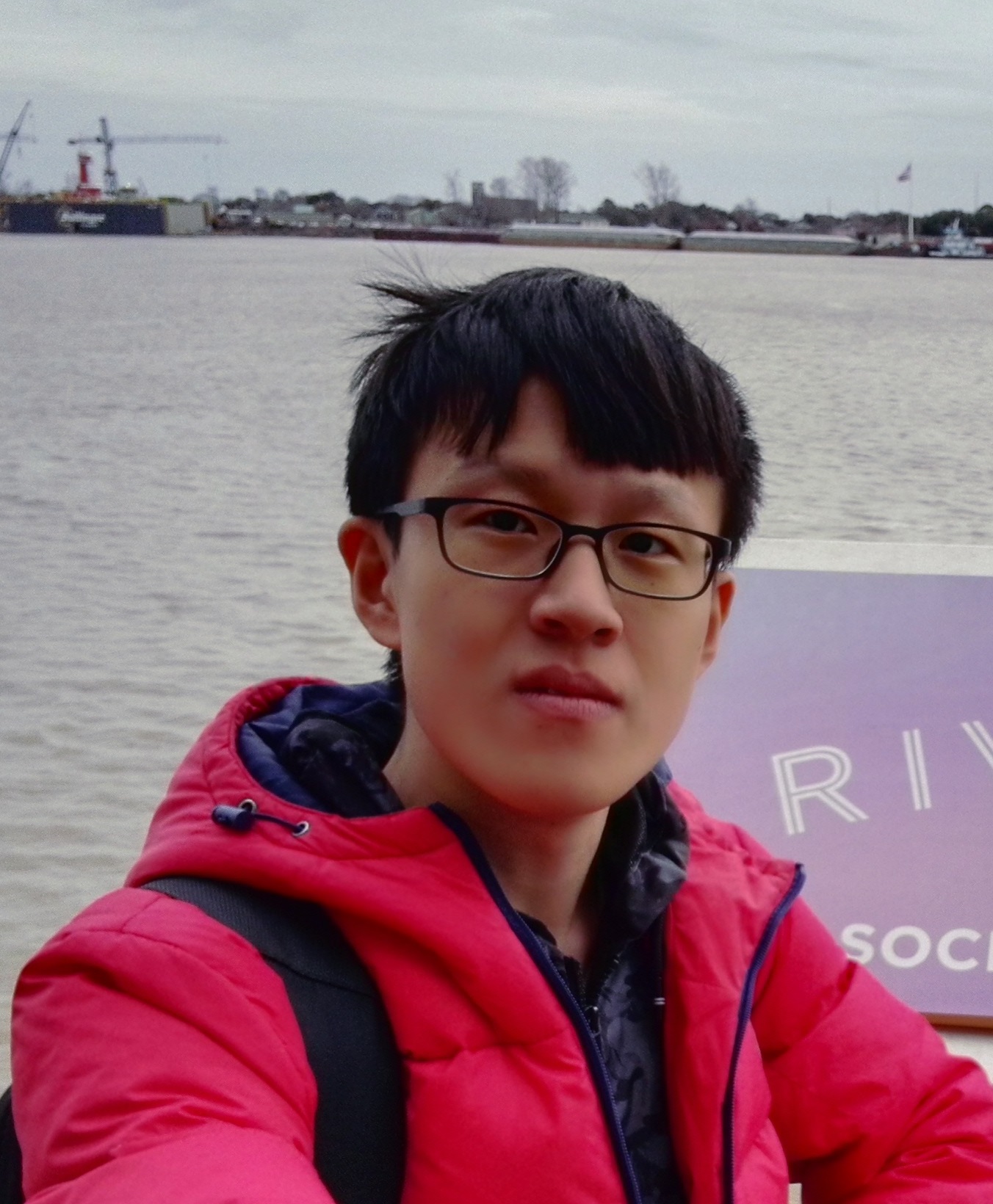}}]{Xiaowei Hu}
	
	received the B.Eng. degree in the Computer Science and Technology from South China University of Technology, China, in 2016. He is currently working toward the Ph.D. degree with the Department of Computer Science and Engineering, The Chinese University of Hong Kong. His research interests include computer vision and deep learning. Mr. Hu is a recipient of the Hong Kong Ph.D. Fellowship.
	
\end{IEEEbiography}

\vspace{-5mm}

\begin{IEEEbiography}[{\includegraphics[width=1in,height=1.25in,clip,keepaspectratio]{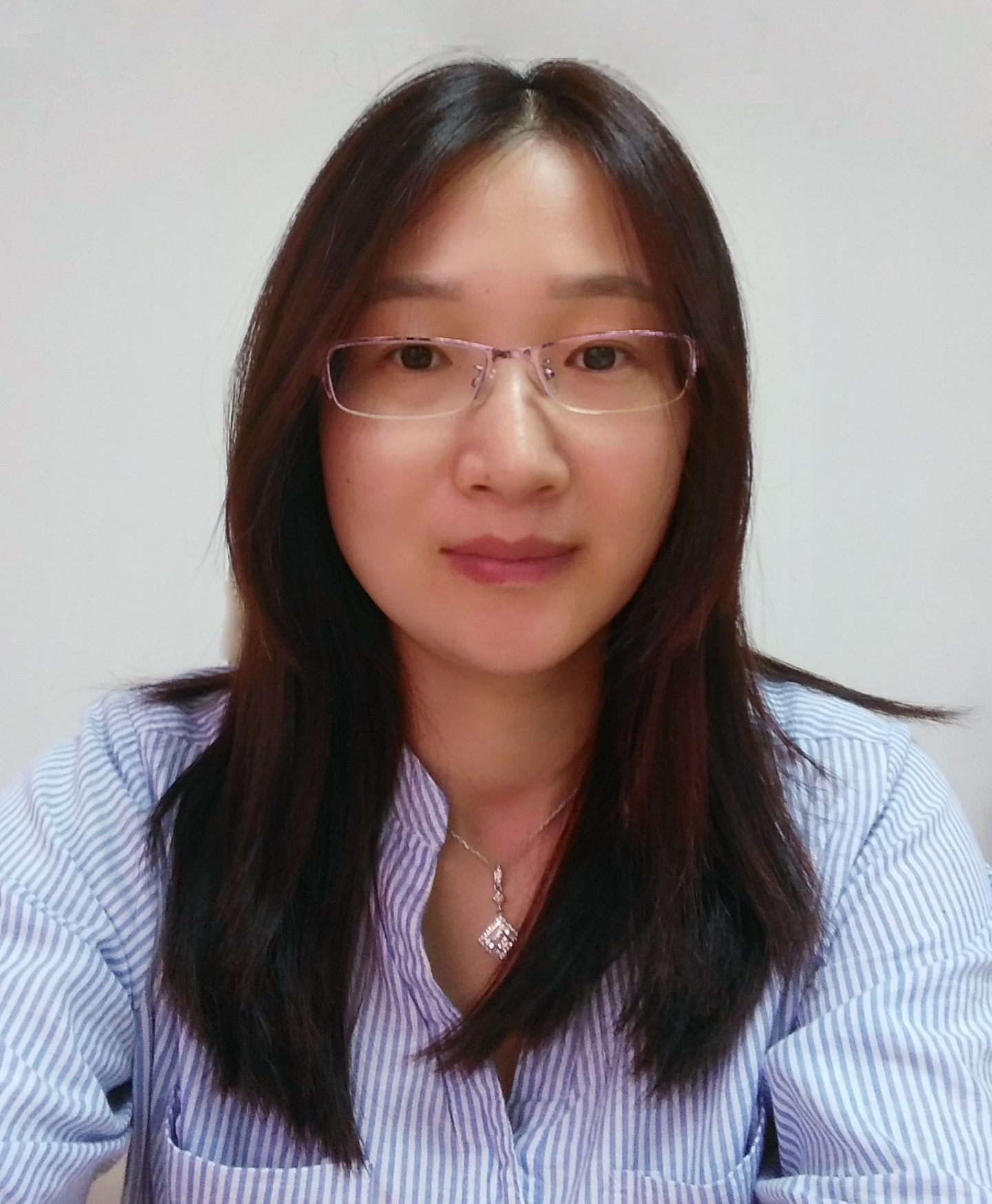}}]{Xuemiao Xu}
	
	received her B.S. and M.S. degrees in Computer Science and Engineering from South China University of Technology in 2002 and 2005 respectively, and Ph.D. degree in Computer Science and Engineering from The Chinese University of Hong Kong in 2009. She is currently a professor in the School of Computer Science and Engineering, South China University of Technology. Her research interests include object detection, tracking, recognition, and image, video understanding and synthesis, particularly their applications in the intelligent transportation system.

\end{IEEEbiography}

\vspace{-5mm}

\begin{IEEEbiography}[{\includegraphics[width=1in,height=1.25in,clip,keepaspectratio]{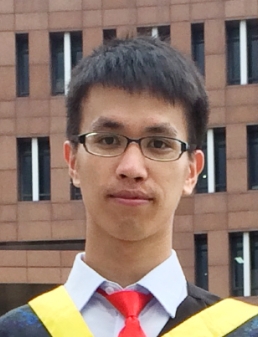}}]{Yongjie Xiao}
	
	received the B.Eng. degree in the Computer Science and Technology from South China University of Technology, China, in 2016. He is currently working toward the Master degree with the School of Computer Science and Engineering, South China University of Technology. His research interests include intelligent transportation, object detection and deep learning.

\end{IEEEbiography}

\vspace{-5mm}

\begin{IEEEbiography}[{\includegraphics[width=1in,height=1.25in,clip,keepaspectratio]{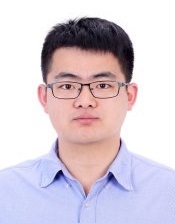}}]{Hao Chen}received the Ph.D. degree in Computer Science and Engineering from The Chinese University of Hong Kong, China, in 2017. He is currently a post-doctoral fellow in The Chinese University of Hong Kong. His research interests include medical image analysis, deep learning, and health informatics. Dr. Chen was a recipient of the Hong Kong Ph.D. Fellowship.
	
\end{IEEEbiography}

\vspace{-5mm}

\begin{IEEEbiography}[{\includegraphics[width=1in,height=1.25in,clip,keepaspectratio]{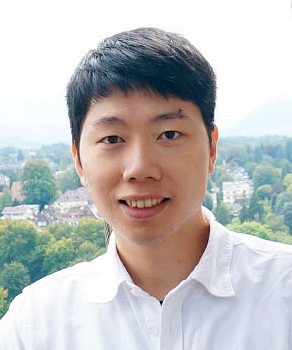}}]{Shengfeng He} obtained his B.Sc. degree and M.Sc. degree from Macau University of Science and Technology and his Ph.D. degree from City University of Hong Kong. He is an Associate Professor in the School of Computer Science and Engineering at South China University of Technology. He was a Research Fellow at City University of Hong Kong and a visiting Ph.D. student at Georgia Institute of Technology. His research interests include computer vision, image processing, computer graphics, and deep learning.
\end{IEEEbiography}

\vspace{-5mm}

\begin{IEEEbiography}[{\includegraphics[width=1in,height=1.25in,clip,keepaspectratio]{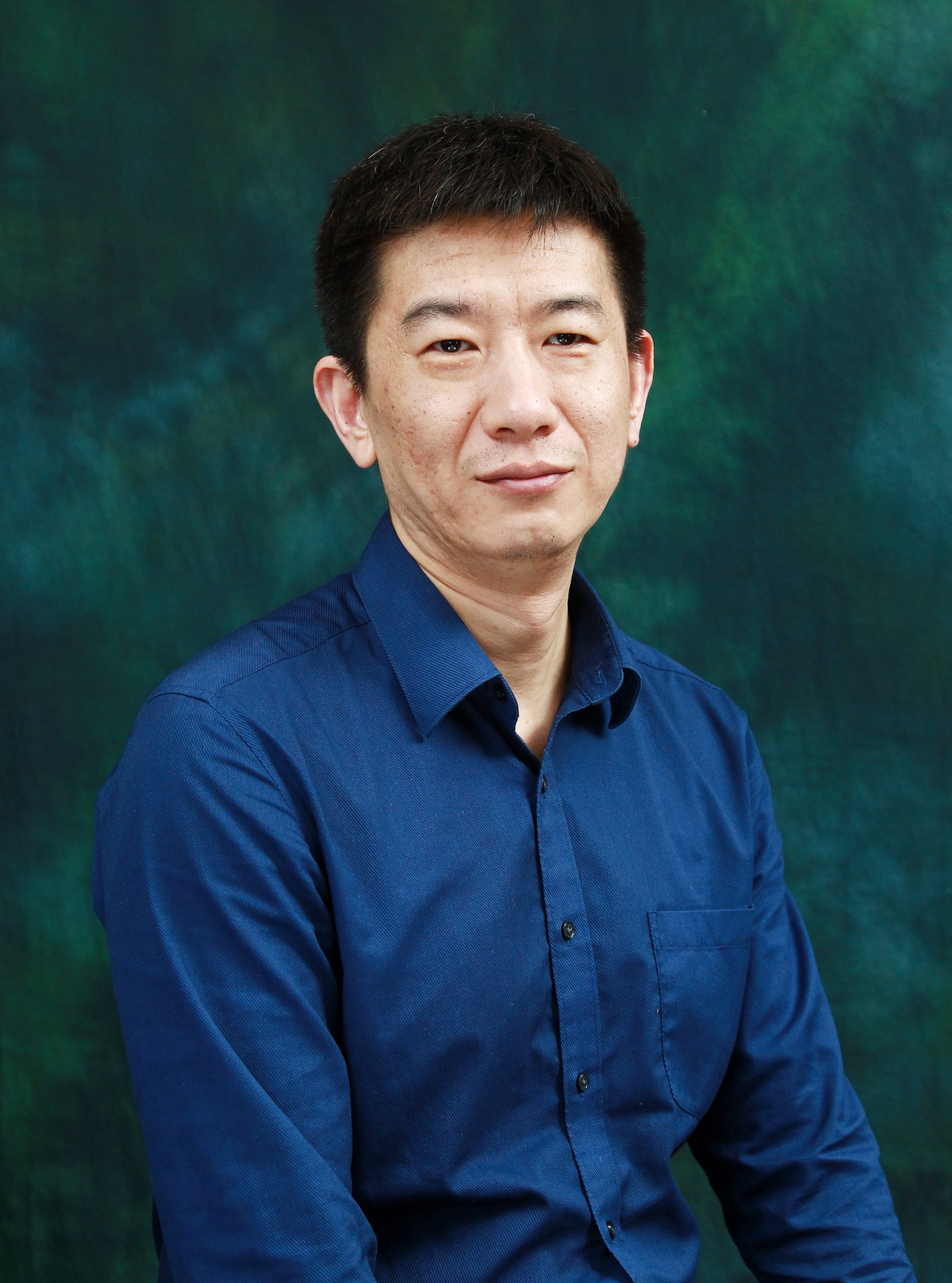}}]{Jing Qin} received his Ph.D. degree in Computer Science and Engineering from the Chinese University of Hong Kong in 2009. He is currently an assistant professor in School of Nursing, The Hong Kong Polytechnic University. He is also a key member in the Centre for Smart Health, SN, PolyU, HK. His research interests include innovations for healthcare and medicine applications, medical image processing, deep learning, visualization and human-computer interaction and health informatics.
\end{IEEEbiography}

\vspace{-5mm}

\begin{IEEEbiography}[{\includegraphics[width=1in,height=1.25in,clip,keepaspectratio]{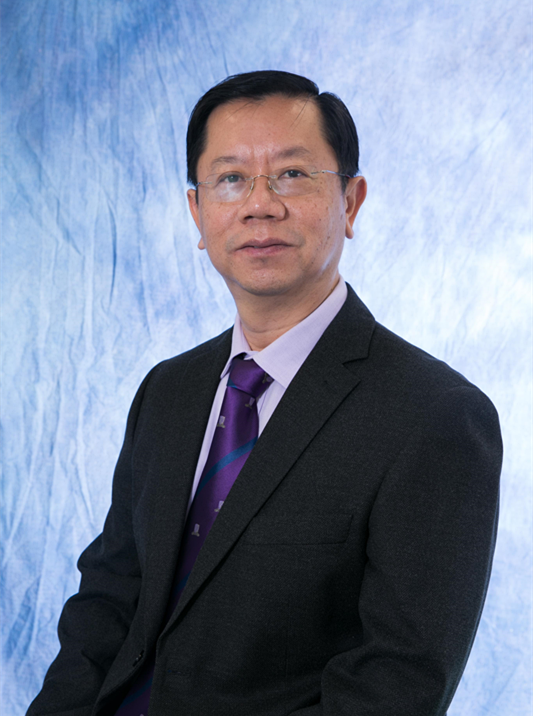}}]{Pheng-Ann Heng} received his B.Sc. from the National University of Singapore in 1985. He received his MSc (Comp. Science), M. Art (Applied Math) and Ph. D (Comp. Science) all from the Indiana University of USA in 1987, 1988, 1992 respectively. He is a professor at the Department of Computer Science and Engineering at The Chinese University of Hong Kong (CUHK). He has served as the Director of Virtual Reality, Visualization and Imaging Research Center at CUHK since 1999 and as the Director of Center for Human-Computer Interaction at Shenzhen Institute of Advanced Integration Technology, Chinese Academy of Science/CUHK since 2006. He has been appointed as a visiting professor at the Institute of Computing Technology, Chinese Academy of Sciences as well as a Cheung Kong Scholar Chair Professor by Ministry of Education and University of Electronic Science and Technology of China since 2007. His research interests include AI and VR for medical applications, surgical simulation, visualization, graphics and human-computer interaction.
\end{IEEEbiography}

\end{document}